\documentclass[11pt]{article}  

\usepackage[final]{automl}

\usepackage{microtype} 
\usepackage{booktabs}  
\usepackage{url}  
\usepackage{csquotes}
\usepackage{natbib}
\usepackage{xcolor}

\usepackage[utf8]{inputenc}
\usepackage{url}

\usepackage{amsmath}
\usepackage{bbm}
\usepackage{bm}
\usepackage{cancel}
\usepackage{mathtools}

\usepackage{ascmac}  
\usepackage{algorithm}
\usepackage{algpseudocode}

\usepackage{fancyhdr}
\usepackage{tcolorbox}
\usepackage{tikz}

\usepackage{booktabs}
\usepackage{comment}
\usepackage{lastpage}
\usepackage{listings}
\usepackage{multibib}
\usepackage{multirow}
\usepackage[super]{nth}

\tcbuselibrary{breakable, skins, theorems}

\allowdisplaybreaks

\newtheorem{definition}{Definition}

\algtext*{EndFor}
\algtext*{EndWhile}
\algtext*{EndIf}
\algtext*{EndProcedure}
\algtext*{EndFunction}
\algnewcommand{\LineComment}[1]{\State \(\triangleright\) #1}

\newcites{appx}{References}

\newif\ifunderreview
\newif\ifsubmission
\newif\ifappendix

\underreviewfalse

\submissiontrue

\appendixtrue

\ifsubmission
\newcommand{\todo}[1]{}
\newcommand{\replace}[2]{}
\newcommand{\sw}[1]{}

\else
\newcommand{\todo}[1]{\textbf{\textcolor{red}{[TODO: #1]}}}

\newcommand{\replace}[2]{\textbf{\textcolor{red}{[del: \cancel{#1}]}}\textbf{\textcolor{blue}{[new: #2]}}}

\newcommand{\sw}[1]{\textbf{\textcolor{cyan}{[SW: #1]}}}

\usepackage[inline]{showlabels}

\fi

\makeatletter
\newcommand{\customlabel}[2]{%
   \protected@write \@auxout {}{\string \newlabel {#1}{{#2}{\thepage}{#2}{#1}{}} }%
   \hypertarget{#1}{}
}
\makeatother

\ifunderreview
  \hypersetup{%
    pdfauthor={Paper under double-blind review}, 
    pdftitle={Python Tool for Visualizing Variability of Pareto Fronts over Multiple Runs},
    pdfsubject={EAS toolkit paper},
    pdfkeywords={Python tool, Multi-objective optimization, Interpretation}
  }
\else
  \hypersetup{%
    pdfauthor={Shuhei Watanabe}, 
    pdftitle={Python Tool for Visualizing Variability of Pareto Fronts over Multiple Runs},
    pdfsubject={EAS toolkit paper},
    pdfkeywords={Python tool, Multi-objective optimization, Interpretation}
  }
\fi

\newcommand{\indic}[1]{\mathbb{I}[#1]}

\newcommand{\ceil}[1]{\lceil #1 \rceil}

\renewcommand{\eqref}[1]{Eq.~(\ref{#1})}

\newcommand{\fv}{\boldsymbol{f}}
\newcommand{\rv}{\boldsymbol{r}}
\newcommand{\xv}{\boldsymbol{x}}
\newcommand{\yv}{\boldsymbol{y}}
\newcommand{\D}{\mathcal{D}}
\newcommand{\F}{\mathcal{F}}
\newcommand{\Pcal}{\mathcal{P}}
\newcommand{\Scal}{\mathcal{S}}
\newcommand{\X}{\mathcal{X}}
\newcommand{\Y}{\mathcal{Y}}

\DeclareFixedFont{\ttb}{T1}{txtt}{bx}{n}{12} 
\DeclareFixedFont{\ttm}{T1}{txtt}{m}{n}{12}  

\definecolor{deepblue}{rgb}{0,0,0.5}
\definecolor{deepred}{rgb}{0.6,0,0}
\definecolor{deepgreen}{rgb}{0,0.5,0}

\begin{document}
\title{Python Tool for Visualizing Variability of \\ Pareto Fronts over Multiple Runs}

\ifunderreview
\author{
  Paper under double-blind review \\
}
\else
\author{
  Shuhei Watanabe \\
  Department of Computer Science, University of Freiburg, Germany\\
  \texttt{watanabs@cs.uni-freiburg.de}
}
\fi

\maketitle

\begin{abstract}
  Hyperparameter optimization is crucial to achieving high performance in deep learning.
  On top of the performance, other criteria such as inference time or memory requirement often need to be optimized due to some practical reasons.
  This motivates research on multi-objective optimization (MOO).
  However, Pareto fronts of MOO methods are often shown without considering the variability caused by random seeds and this makes the performance stability evaluation difficult.
  Although there is a concept named \emph{empirical attainment surface} to enable the visualization with uncertainty over multiple runs, there is no major Python package for empirical attainment surface.
  We, therefore, develop a Python package for this purpose and describe the usage.
  The package is available at \url{https://github.com/nabenabe0928/empirical-attainment-func}.
  OptunaHub~(\cite{ozaki2026optunahub}) now provides our implementation~\footnote{\url{https://hub.optuna.org/visualization/plot_empirical_attainment_surface/}}.
\end{abstract}

\section{Introduction}
Hyperparameter (HP) optimization is an essential step for strong performance in deep learning~(\cite{chen2018bayesian,henderson-aaai18a}).
On top of the performance, other criteria such as inference time or memory requirement often need to be optimized depending on the computational resources available.
This necessitates multi-objective optimization (MOO) and many researchers, therefore, actively work on this problem setting~(\cite{deb2002fast,ozaki2020multiobjective}).

Although MOO methods have been actively studied so far, the comparison of MOO methods is often troublesome because the outcome of MOO methods is a set of multiple optimized metrics.
For this reason, researchers often scalarize the multiple metrics into hypervolume (HV; see Appendix~\ref{appx:background:section:hv}) so that they can compare each method based on a single metric or simply pick up a single instance to avoid complications in a figure.
On the other hand, we surely have a method to retain the uncertainty visualization in Pareto front (PF) via \emph{empirical attainment surface} (EAS)~(\cite{grunert2001inferential}), which is, roughly speaking, a set of good solutions (PF solutions) attained by a certain number of independent runs, and some papers use this method~(\cite{lopez2004design,ozaki2022multiobjective,watanabe2022multi,watanabe2023speeding,awad2023mo}).
The problems of EAS are that (1) the implementation is complicated~\footnote{
  Merging results from multiple runs and obtaining a step-looking surface in Figure~\ref{main:background:fig:pf-conceptual} are complicated enough to make researchers hesitate to implement the method just for small analysis.
} although it is not impossible to implement the method by themselves, and (2) only the R package~\cite{lopez2010exploratory} is available and no Python package exists to the best of our knowledge.

To this end, we developed an accessible EAS visualization package in Python and describe its usage in this paper.
Users can plot EAS simply by providing an array of results from multiple independent runs with an added option for plotting HV over time.
Note that our package computes the exact EAS rather than the discretized approximation proposed by \cite{knowles2005summary} and focuses on the bi-objective case, as interpreting EAS with more than $2$ objectives is hard.
We defer the details of the tri-objective EAS visualization to \cite{tuvsar2013approach}.
For those who would like to use our package for such cases, combining it with multidimensional scaling~(\cite{kruskal1978multidimensional,mead1992review}) might be an option, although it risks distorting the geometric structures that make EAS intuitive to interpret.

\section{Background}

Throughout the paper, we use the notation $[i] \coloneqq \{1,2,\dots, i\}$ where $i$ is a positive integer.

\subsection{Multi-Objective Optimization}
Suppose we have a set of criteria $\{f_m(\xv)\}_{m=1}^M$ to optimize, then MOO is formalized as follows:
\begin{equation}
\begin{aligned}
  \min_{\xv \in \X}\fv(\xv) \coloneqq \min_{\xv \in \X}~[f_1(\xv), f_2(\xv), \dots, f_M(\xv)]
\end{aligned}
\end{equation}
where $\fv: \X \rightarrow \mathbb{R}^{M}$ is the objective function, $\xv \in \X$ is a HP configuration, $\X \coloneqq \X_1 \times \X_2 \times \cdots \times \X_D$ is the search space, and $\X_d \subseteq \mathbb{R}$ is the domain of the $d$-th HP.
For example, the learning rate and the dropout rate of neural networks would be HPs, and validation error and runtime would be the objectives.
Although we discuss the general concepts for MOO, we note that our package only supports the visualization of bi-objective optimization.
Users could apply multidimensional scaling~(\cite{kruskal1978multidimensional,mead1992review}) to the obtained objective vectors and then compute EAS of the reduced objective vectors.
However, multidimensional scaling may not maintain the geometric property of PF that is to be able to recognize dominated observations by checking the geometrical locations of PF and observations.

\subsection{Pareto Front (PF)}
\begin{figure}
  \centering
  \includegraphics[width=0.9\textwidth]{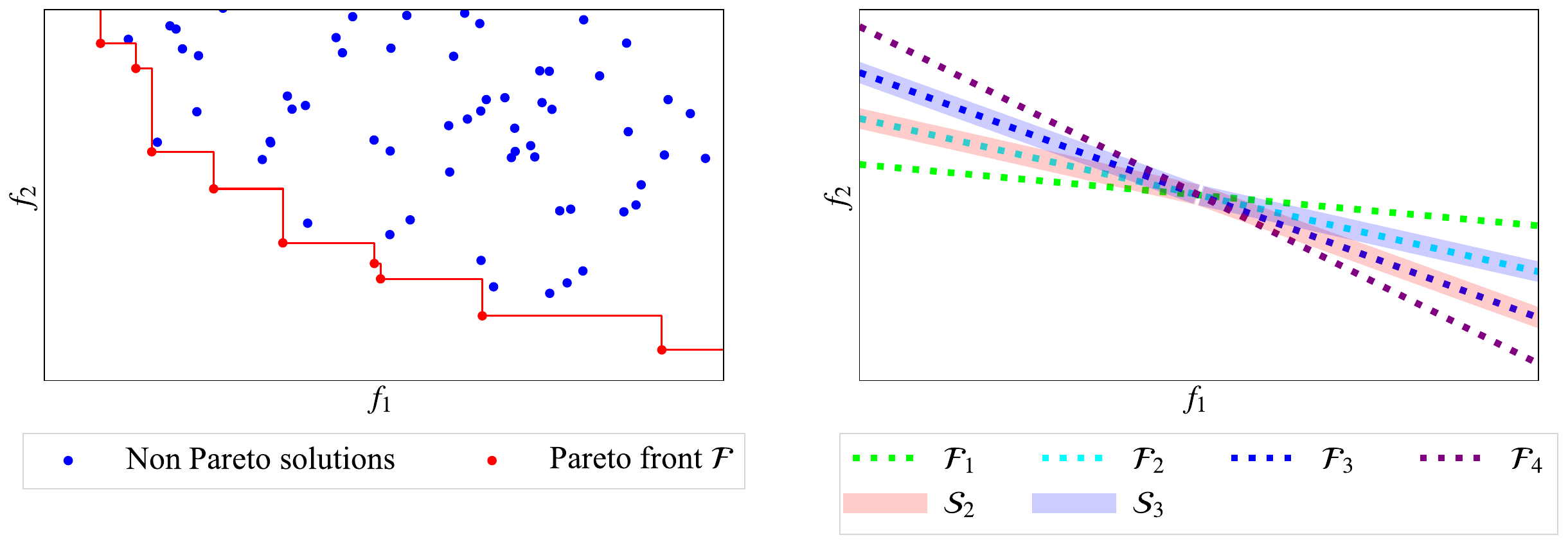}
  \caption{
    Conceptual visualizations of PF (\textbf{Left}) and EASs (\textbf{Right}).
    Lower is better for both objectives $f_1, f_2$.
    \textbf{Left}: an example of PF.
    The red dots in the objective space are the objective vectors that belong to PF $\F$ (the red solid line).
    The blue dots are dominated by the objective vectors in $\F$.
    In other words, the blue dots do not lie on PF.
    \textbf{Right}: an example of EASs.
    Each dotted line is the PF of one of four independent runs.
    Light-colored bands show EASs of $L = 2$ (red) and $L = 3$ (blue).
    The red surface is attained by PFs from two independent runs, and the blue surface is attained by PFs from three independent runs.
  }
  \label{main:background:fig:pf-conceptual}
\end{figure}

Due to the trade-off among multiple objectives, optimal solutions are usually not unique and this necessitates the exact definition of optimal solutions in MOO.
Given a set of observations $\D \coloneqq \{(\xv_n, \fv(\xv_n))\}_{n=1}^{N}$, we first define dominance.
\begin{definition}[Dominance]
  Given two vectors $\yv_i, \yv_j \in \mathbb{R}^M$ in the objective space, $\yv_i$ is said to weakly dominate $\yv_j$ if and only if $y_{i,m} \leq y_{j,m}$ holds for all $m \in [M]$, where $y_{i,m}, y_{j,m} \in \mathbb{R}$ are the $m$-th elements of $\yv_i$ and $\yv_j$, respectively.
  We write this relationship as $\yv_i \preceq \yv_j$.
  Furthermore, $\yv_i$ is said to dominate $\yv_j$ if and only if $\yv_i \preceq \yv_j$ holds and there exists $m \in [M]$ such that $y_{i,m} < y_{j,m}$.
  We write this relationship as $\yv_i \prec \yv_j$.
\end{definition}
Next, we define dominance and weak dominance between a set of vectors and a vector:
\begin{definition}[Dominance between a set of vectors and a vector]
  Given a set of objective vectors $Y \coloneqq \{\yv_n\}_{n=1}^N$ and an objective vector $\yv$, $Y$ is said to dominate (or weakly dominate) $\yv$ if and only if there exists $n \in [N]$ such that $\yv_n \prec \yv$ (resp. $\yv_n \preceq \yv$).
  We write this relationship as $Y \prec \yv$ (resp. $Y \preceq \yv$).
\end{definition}
Let $\Y \subseteq \mathbb{R}^M$ be the objective space.
The PF of a set of observations $\D$ is then defined as follows:
\begin{definition}[Pareto front (PF)]
  Given a set of observations $\D \coloneqq \{(\xv_n, \fv(\xv_n))\}_{n=1}^N$, a set of PF solutions is defined as $\Pcal \coloneqq \{\xv_n \mid \nexists n^\prime \in [N], \fv(\xv_{n^\prime}) \prec \fv(\xv_n)\}$.
  Then PF is a set of non-dominated objective vectors defined as $\F \coloneqq \{\yv \in \Y \mid \forall \xv \in \Pcal, \fv(\xv) \nprec \yv, \yv \nprec \fv(\xv)\}$.
\end{definition}
See Figure~\ref{main:background:fig:pf-conceptual} for the intuition.
Note that PF requires the time complexity of $O(N(\log N)^{M - 2})$ for $M > 3$ and $O(N\log N)$ for $M = 2, 3$ by Kung's algorithm~(\cite{kung1975finding}) and since a major Python package \texttt{pygmo} does not support three or more objectives, practitioners might want to use
\ifunderreview
a minor package named \texttt{fast-pareto}.
\else
our package at \url{https://github.com/nabenabe0928/fast-pareto/}.
\fi
In principle, the ultimate goal of MOO is to collect all possible PF solutions.
Therefore, it is crucial to visualize PF to check whether the obtained solutions cover PF densely.
We would like to validate the diversity in PF when researchers develop an MOO method.
However, PF of a single run does not take the uncertainty caused by random seeds into account, and thus we need to consider the uncertainty by using EAS~(\cite{knowles2005summary,fonseca2011computation}).
We describe the method in the next subsection.

\subsection{Empirical Attainment Surface (EAS)}
Suppose we have obtained a set of PFs $\{\F_s\}_{s=1}^S$ from $S$ independent runs, each with a different random seed.
EAS is defined by the following empirical attainment function:
\begin{equation}
\begin{aligned}
  \alpha(\yv) \coloneqq \alpha(\yv \mid \F_1, \dots, \F_S) = \frac{1}{S}\sum_{s=1}^S \indic{\F_s \preceq \yv}.
\end{aligned}
\end{equation}
Intuitively speaking, if $\alpha(\yv) = k/S$ for an integer $k \leq S$, then $\yv$ is weakly dominated by the PFs $\F_s$ of exactly $k$ different runs.
Then $p\%$-EAS is defined as $\Scal \coloneqq \{\yv \in \Y \mid \alpha(\yv) \geq p/100\}$.
Since the percentages can only take values in $\{100 \times s/S\}_{s=0}^S$, EAS in the form of $\Scal_L \coloneqq \{\yv \in \Y \mid \alpha(\yv) \geq L/S\}$ with level $L \in [S]$ is used in practice.
For example, $\Scal_{\ceil{S/2}}$ is the PF achieved by at least half of the independent runs and $\Scal_{\ceil{S/4}}$ is the PF achieved by at least a quarter of the independent runs.
Using two different EASs, we can visualize the uncertainty of PF caused by random seeds.
See Figure~\ref{main:background:fig:pf-conceptual} for the intuition.
If we have only a single objective, $\Scal_{\ceil{S/2}}$ has a single value, which is the median performance over the $S$ independent runs.

\section{Tool Usage}

Our package is available at
\ifunderreview
\url{https://anonymous.4open.science/r/empirical-attainment-func-3346/}.
\else
\url{https://github.com/nabenabe0928/empirical-attainment-func} and the package can be simply installed by \texttt{pip install empirical-attainment-func}.
\fi
Throughout this section, we describe the usage of our package and use the following common code block:

\vspace*{3mm}
\begin{minipage}{0.95\linewidth}
\begin{lstlisting}
rng = np.random.RandomState(0)
n_runs, n_samples, dim = 50, 20, 3
X = rng.random((n_runs, n_samples, dim)) * 10 - 5
# costs.shape = (n_runs, n_samples, 2)
# func is defined in Appendix C
costs = func(X)
\end{lstlisting}
\end{minipage}
Note that \texttt{n\_runs}, \texttt{n\_samples}, and \texttt{dim} correspond to $S$, $N$, and $D$, respectively, and we assume that each code block already imports \texttt{numpy} as \texttt{np} and \texttt{matplotlib.pyplot} as \texttt{plt}, and ends with \texttt{ax.legend()} and \texttt{plt.show()} for plotting.

\subsection{Points on Empirical Attainment Surface}
\label{main:usage:section:eas}

The \texttt{get\_empirical\_attainment\_surface} function computes EASs of the specified levels $L_1, L_2, \dots, L_K \in [S]$.
This function takes the following arguments:
\begin{itemize}
  \vspace{-1mm}
  \item \texttt{costs} (\texttt{np.ndarray}): an array of shape $(S, N, 2)$ containing the objective vectors from independent runs $\{\{\fv(\xv_{s,n})\}_{n=1}^N\}_{s=1}^S$ where $\xv_{s,n}$ is the $n$-th HP in the $s$-th run,
  \vspace{-2mm}
  \item \texttt{levels} (\texttt{List[int]}): a list of levels $[L_1, L_2, \dots, L_K]$. Using $[\lfloor S/4 \rfloor, \lfloor S/2 \rfloor, \lfloor 3S/4 \rfloor]$ yields $[\Scal_{\lfloor S/4 \rfloor}, \Scal_{\lfloor S/2 \rfloor}, \Scal_{\lfloor 3S/4 \rfloor}]$,
  \vspace{-2mm}
  \item \texttt{larger\_is\_better\_objectives} (\texttt{Optional[List[int]]}): a list of integers to specify the objectives to be maximized (The default setting for each objective is minimization). For example, if we pass the list \texttt{[1]}, the second objective is recognized as an objective to be maximized, and
  \vspace{-2mm}
  \item \texttt{log\_scale} (\texttt{Optional[List[int]]}): a list of integers to specify the objectives to be handled in the log scale. For example, if we pass the list \texttt{[1]}, the second objective is handled in the log scale.
  \vspace{-1mm}
\end{itemize}
An example usage is as follows:

\begin{minipage}{0.95\linewidth}
\begin{lstlisting}[caption=An example of \texttt{get\_empirical\_attainment\_surface}.\label{main:usage:code:get-eas}]
from eaf import get_empirical_attainment_surface
  
levels = [n_runs // 4, n_runs // 2, 3 * n_runs // 4]
# The arguments and output are explained in text
surfs = get_empirical_attainment_surface(costs, levels)
\end{lstlisting}
\end{minipage}
The output is, in principle, $[\Scal_{L_1}, \Scal_{L_2}, \dots, \Scal_{L_K}]$, but it is slightly different to handle the visualization easier.
We first internally compute the union of PF solutions $P \coloneqq \cup_{s=1}^S \Pcal_s$ over all runs and collect all unique values in the first objective $\mathcal{U}_1 = \{f_1(\xv) \mid \xv \in P\} \cup \{\infty, -\infty\}$.
Each $\Scal_{L_k}$ for $k \in [K]$ has shape $(|\mathcal{U}_1|, 2)$, so the returned array has shape $(K, |\mathcal{U}_1|, 2)$, and the first element $y_1$ in $[y_1, y_2] \coloneqq \yv \in \Scal_{L_k}$ always takes a value in $\mathcal{U}_1$.
As the shapes of each $\Scal_{L_k}$ do not match each other without modifications, we pad the same value at the head of each $\Scal_{L_k}$ with an edge point of $\Scal_{L_k}$ so that the returned EAS does not include any objective vectors not on $\Scal_{L_k}$.
By connecting nodes in $\Scal_{L_k}$, we can visualize EAS with the level of $L_k$.
Note that $\Scal_{L_k}$ is sorted by $y_1$ in $[y_1, y_2] \coloneqq \yv \in \Scal_{L_k}$ by default.

\begin{figure}
  \centering
  \includegraphics[width=0.9\textwidth]{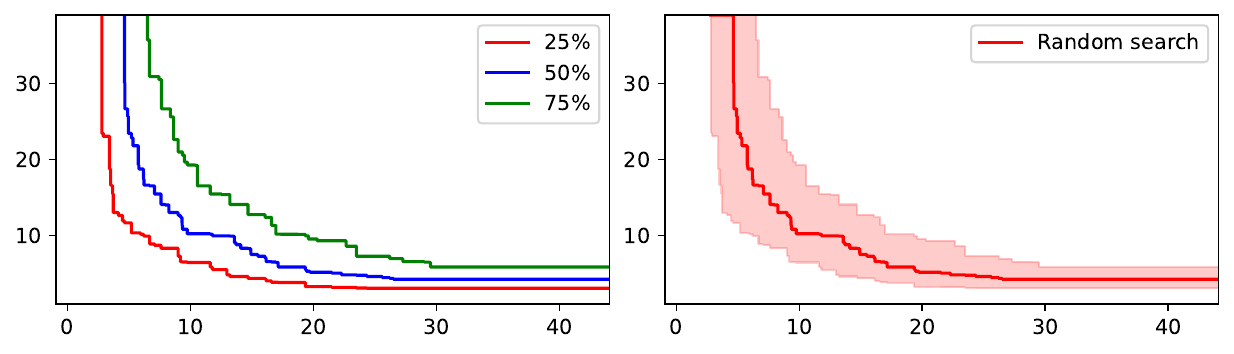}
  \caption{
    Visualizations generated by Listing~\ref*{main:usage:code:plot-surface} (\textbf{Left}) and Listing~\ref*{main:usage:code:plot-surface-with-band} (\textbf{Right}).
    The horizontal axis shows $f_1$ and the vertical axis shows $f_2$.
    \textbf{Left}: an example plot of EASs. By repeating the same process for different MOO methods, we can compare multiple methods.
    \textbf{Right}: an example plot of EAS with an uncertainty band.
    The light-colored band shows the variability of PF due to random seeds.
  }
  \label{main:usage:fig:plot-surfaces}
\end{figure}

\subsection{Plot of Empirical Attainment Surfaces}
For plots, we use \texttt{EmpiricalAttainmentFuncPlot} and it receives the following arguments for internal processing:
\begin{itemize}
  \vspace{-1mm}
  \item \texttt{larger\_is\_better\_objectives} (\texttt{list[int] | None}): Same as in Section~\ref{main:usage:section:eas},
  \vspace{-2mm}
  \item \texttt{log\_scale} (\texttt{list[int] | None}): Same as in Section~\ref{main:usage:section:eas},
  \vspace{-2mm}
  \item \texttt{true\_pareto\_sols} (\texttt{np.ndarray | None}): an array of the true PF if available. It is required if we plot normalized HV (see Appendix~\ref{appx:usage:section:plot-hv}),
  \vspace{-2mm}
  \item \texttt{x\_min}, \texttt{x\_max}, \texttt{y\_min}, \texttt{y\_max} (\texttt{float | None}): the lower/upper bounds of the first/second objective if available. These values could be used to fix the value ranges of each objective, and
  \vspace{-2mm}
  \item \texttt{ref\_point} (\texttt{np.ndarray | None}): the reference point of the problem of interest if available. It is required if we plot HV.
  \vspace{-1mm}
\end{itemize}
Although all of them are optional, additional information could enable extra plots such as HV and the surface of PF.
Assuming that we already computed \texttt{surfs} ($[\Scal_{\lfloor S/4 \rfloor}, \Scal_{\lfloor S/2 \rfloor}, \Scal_{\lfloor 3S/4 \rfloor}]$) using Listing~\ref{main:usage:code:get-eas}, then we can connect the nodes in $\Scal_{L_k}$ simply by the following:

\begin{minipage}{0.95\linewidth}
\begin{lstlisting}[caption=An example of \texttt{plot\_multiple\_surface}.\label{main:usage:code:plot-surface}]
from eaf import EmpiricalAttainmentFuncPlot
  
_, ax = plt.subplots()
eaf_plot = EmpiricalAttainmentFuncPlot()
colors = ["red", "blue", "green"]
labels = ["25%", "50%", "75%"]
eaf_plot.plot_multiple_surface(
  ax, colors=colors, labels=labels, surfs=surfs
)
\end{lstlisting}
\end{minipage}

The figure generated by this code is available in Figure~\ref{main:usage:fig:plot-surfaces} (\textbf{Left}).
The arguments \texttt{ax}, \texttt{surfs}, \texttt{colors}, and \texttt{labels} are required, and \texttt{linestyles}, \texttt{markers}, and \texttt{kwargs} are optional and forwarded to \texttt{ax.plot}.
The $k$-th element of each list is passed to \texttt{ax.plot} for the $k$-th surface.
If we need to plot only a single EAS, \texttt{eaf\_plot.plot\_surface} could be used instead.
When we would like to visualize the uncertainty band of PF, \texttt{plot\_surface\_with\_band} or \texttt{plot\_multiple\_surface\_with\_band} could be used:

\begin{minipage}{0.95\linewidth}
\begin{lstlisting}[caption=An example of \texttt{plot\_surface\_with\_band}.\label{main:usage:code:plot-surface-with-band}]
eaf_plot.plot_surface_with_band(
    ax, color="red", label="Random search", surfs=surfs
)
\end{lstlisting}
\end{minipage}

Figure~\ref{main:usage:fig:plot-surfaces} (\textbf{Right}) visualizes the result obtained by this code.
In Figure~\ref{main:usage:fig:plot-surfaces} (\textbf{Left}), the red line corresponds to $\Scal_{L_1}$, the blue line corresponds to $\Scal_{L_2}$, and the green line corresponds to $\Scal_{L_3}$.
We can easily plot EASs of various MOO methods using \texttt{plot\_multiple\_surface\_with\_band} as well.
The band is determined by $\Scal_{L_1}, \Scal_{L_3}$ and the center line is $\Scal_{L_2}$.
Note that \texttt{surfs} for \texttt{plot\_surface\_with\_band} and each element in \texttt{surfs\_list} for \texttt{plot\_multiple\_surface\_with\_band} must be an array with shape $(K, |\mathcal{U}_1|, 2)$ where $K = 3$, obtained from Listing~\ref{main:usage:code:get-eas}.

\section{Conclusions}
In this paper, we introduced a Python package to visualize uncertainty bands for MOO.
As described, our package is very easy to use and users can easily get the observations on EAS and EAS itself.
Although recent research papers often focus on the visualization of HV over time or PF from a single run, this easy-to-use package allows researchers to qualitatively evaluate the uncertainty in their MOO methods and potential readers would not be left behind due to the lack of interpretability caused by the information loss either from HV or choosing a single random seed.
Broad impact \& limitations are described in Appendix~\ref{appx:broad-impact:section}.

\clearpage

\bibliographystyle{apalike}
\bibliography{ref}

\ifappendix
\clearpage
\appendix
\section{Hypervolume (HV)}
\label{appx:background:section:hv}
HV is a widely used metric to compare the performance of various MOO methods.
To compute HV of a set of observations $\D$, we first define a \emph{reference point} $\rv \in \mathbb{R}^M$ so that the reference point is dominated by all the observations in $\D$.
More formally, the reference point is chosen from $\{\yv \in \mathbb{R}^M \mid \fv(\xv_n) \prec \yv \ \forall n \in [N]\}$.
In practice, if users already know the possible worst (in our case maximum) value for each objective, these values will be used.
Then HV is defined as follows:
\begin{definition}[Hypervolume (HV)]
  Let $\mu$ be the Lebesgue measure defined on $\mathbb{R}^M$ and $\F$ be the PF in the objective space $\Y$.
  Then HV is computed as:
  \begin{equation}
  \begin{aligned}
    V(\F | \rv) \coloneqq \mu(\{\yv \in \mathbb{R}^M \mid \F \prec \yv \prec \rv \}).
  \end{aligned}
  \end{equation}
\end{definition}
Roughly speaking, the Lebesgue measure quantifies the volume of a set.
Note that rescaling each objective's range into $[0, 1]$, i.e., normalization, only multiplies HV by a constant factor and thus does not change the ranking of methods, and normalized HV is used in practice if we know the best possible PF.

Although HV is a convenient metric for comparison, a high HV does not guarantee that MOO methods yielded a diverse set of solutions.
For example, let the reference point be $\rv = [1.1, 1.1]$ and $\fv = (f_1, f_2)$ be a mapping $\X \rightarrow [0, 1] \times [0, 1]$.
If $f_1$ is uniformly distributed in $0 \leq f_1 \leq 10^{-5}$ while $f_2$ is uniformly distributed in $[0,1]$ in the objective space, although $f_1 \leq 10^{-7}$ is the top-$1\%$ solution, HV changes by a factor of only $(1.1 - 10^{-7}) / (1.1 - 10^{-5})\simeq 1$ in comparison to a set of trivial solutions.
On the other hand, when we yield the top-$1\%$ solution in $f_2$, which is $0.01$, HV changes by a factor of $(1.1 - 0.01) / (1.1 - 1) = 10.9$.
In other words, one can inflate HV by focusing on the optimization of $f_2$ alone.
From this example, we can recognize the potential problems that HV might cause due to its heavy dependency on the choice of the reference point $\rv$ and the distribution of the objective vector $p(\yv)$.
This example also implies that the reference point should be tight and that the standard error taken over multiple independent runs can be extremely small depending on the distribution $p(\yv)$.

\begin{figure}
  \centering
  \includegraphics[width=0.6\textwidth]{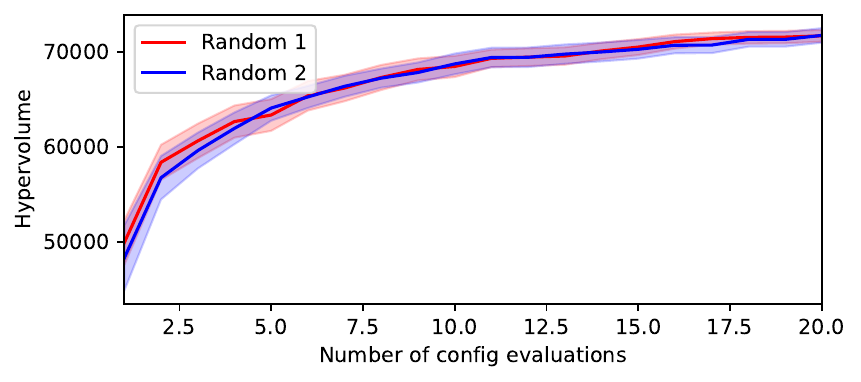}
  \caption{
    The visualization obtained by Listing~\ref{appx:usage:code:plot-hv}.
    This example plots HV over time with a band showing the standard error.
  }
  \label{appx:usage:fig:plot-hv}
\end{figure}

\section{Hypervolume Plot over Time}
\label{appx:usage:section:plot-hv}
Our package also supports the plot of HV for $M = 2$ over time and the visualization obtained by Listing~\ref{appx:usage:code:plot-hv} is provided in Figure~\ref{appx:usage:fig:plot-hv}.

\begin{minipage}{0.95\linewidth}
\begin{lstlisting}[caption=An example of \texttt{plot\_multiple\_hypervolume2d\_with\_band}.\label{appx:usage:code:plot-hv}]
X2 = rng.random((n_runs, n_samples, dim)) * 10 - 5
costs2 = func(X2)
stacked_costs = np.stack([costs, costs2])
  
ref_point = np.array([75, 1029])  # problem specific
eaf_plot = EmpiricalAttainmentFuncPlot(
    ref_point=ref_point
)
colors = ["red", "blue"]
labels = ["Random 1", "Random 2"]
  
eaf_plot.plot_multiple_hypervolume2d_with_band(
    ax,
    costs_array=stacked_costs,
    colors=colors,
    labels=labels,
    normalize=False,
)
\end{lstlisting}
\end{minipage}

We can use the \texttt{normalize} option only if we have the true set of PF.

\section{Objective Function Used in the Examples}
\label{appx:usage:section:func}
\begin{minipage}{0.95\linewidth}
\begin{lstlisting}[caption=The definition of the objective function used in the examples.\label{appx:usage:code:func}]
def func(x: np.ndarray) -> np.ndarray:
    f1 = np.sum(x**2, axis=-1)
    f2 = np.sum((x - 2)**2, axis=-1)
    costs = np.stack([f1, f2], axis=-1)
    return costs
\end{lstlisting}
\end{minipage}

\section{Broad Impact \& Limitations}
\label{appx:broad-impact:section}
Our package allows researchers to easily provide PFs with uncertainty bands and it would benefit research papers in terms of the transparency in the results' interpretation.
The primary limitation of our package is the lack of visualization for many-objective optimization.
As described, users could apply multidimensional scaling to the obtained objective vectors so that we can visualize the reduced objective vectors in a 2D plane.
However, our package does not provide such a feature yet.


\else

\customlabel{data1}{1}
\customlabel{data2}{2}

\fi

\end{document}